\documentclass[8pt, twocolumn]{extarticle} 
\usepackage{a4wide}

\usepackage{subcaption}
\usepackage{todonotes}
\usepackage[colorlinks=true,allcolors=blue]{hyperref}%
\usepackage{amsmath}
\usepackage{amsfonts}
\usepackage{adjustbox}
\usepackage{siunitx}
\usepackage{soul} 
\usepackage{enumitem}
\usepackage[super,comma]{natbib}
\usepackage{abstract}
\usepackage[ruled,vlined]{algorithm2e}

\usepackage{xcolor}

\newcommand\hmm[1]{\ifnum\ifhmode\spacefactor\else2000\fi>1000 \uppercase{#1}\else#1\fi} 
\newcommand{\tol}{Triangle of Life}

\usepackage{tcolorbox}
\newtcolorbox{mybox}[1][lightgray!40!white]{sharp corners,colback=#1,colframe=black,
before upper={\rule[-3pt]{0pt}{10pt}},boxrule=1.5pt,
before skip=6pt, after skip= 6pt,
bottom=3pt,left=7pt,right=7pt,top=3pt
}

\title{\huge \bf Learning Locomotion Skills in Evolvable Robots}

\author{
Gongjin Lan$^{1*}$ 
\and Maarten van Hooft$^{1,2}$ 
\and Matteo De Carlo$^1$ 
\and Jakub M. Tomczak$^1$ 
\and A.E. Eiben$^{1}$ \\
$^1$ Department of Computer Science, Vrije Universiteit Amsterdam, Amsterdam, the Netherlands. \\
$^2$ Universiteit van Amsterdam, Amsterdam, the Netherlands. \\
$^*$e-mail: g.lan@vu.nl
}

\date{}
\begin{document}

\twocolumn[
\maketitle
\begin{abstract}
\noindent \textbf{The challenge of robotic reproduction --making of new robots by recombining two existing ones-- has been recently cracked and physically evolving robot systems have come within reach. Here we address the next big hurdle: producing an adequate brain for a newborn robot. In particular, we address the task of targeted locomotion which is arguably a fundamental skill in any practical implementation. We introduce a controller architecture and a generic learning method to allow a modular robot with an arbitrary shape to learn to walk towards a target and follow this target if it moves. Our approach is validated on three robots, a spider, a gecko, and their offspring, in three real-world scenarios.}
\end{abstract}
\vspace{.1cm} 
]


    
\noindent An exciting area where biology meets technology is evolutionary robotics \cite{nolfi2000evolutionary,Vargas-2014,doncieux2015evolutionary}. The key idea behind the field is to optimize the design of robots through evolutionary computing \cite{eiben2003introduction-to}. Using artificial evolution for robot design has a strong rationale.

\begingroup
\addtolength\leftmargini{0.1cm}
\begin{quote}
{\bf As natural evolution has produced successful life forms for practically all possible environmental niches on Earth, it is plausible that artificial evolution can produce specialised robots for various environments and tasks.}
\end{quote}
\endgroup


\noindent The long-term vision foresees a technology where robots for a certain application can be `bred' through iterated selection and reproduction cycles until they satisfy the users' criteria. This approach is not meant to replace classic engineering when designing robots for structured environments with known conditions. However, for complex, unstructured environments with (partially) unknown and possibly changing conditions evolution offers great advantages \cite{howard2019evolving}. A crucial difference between a system of evolving robots and a usual evolutionary algorithm is that the members of the population are not digital objects in a virtual space, but physical artefacts in the real-world. Such a system goes beyond evolutionary computing and implements the Evolution of Things with several new challenges rooted in the physical incarnation \cite{Eiben2012Embodied-Artifi,eiben2015evolutionary}.

Another property distinguishing robot evolution from mainstream evolutionary optimization is that robots have agency, i.e., they are active artefacts that exhibit autonomous behavior. To this end, it is important to note that a robot is a combination of its body (morphology, hardware) and its brain (controller, software), and the behavior depends on both \cite{Louise2011Beyond,weigmann2012does}. This implies that the evolution of robots should concern both the bodies and the brains \cite{pfeifer2007body}. 

Thus, in a full-fledged evolutionary robot system the morphologies as well as the controllers undergo evolution. This is in stark contrast with the current practice. Evolutionary robotics today is mainly concerned with evolving the controllers of simulated robots \cite{radhakrishna2018survey}. Systems where morphologies and controllers of robots evolve simultaneously are rare and they work in simulation \cite{Auerbach_Bongard2014Environmental}. Occasionally, an organism evolved in software is constructed in the real world, using hardware \cite{lipson2000automatic} or `wetware' \cite{Kriegman2020}, but the evolutionary process is simulated. This evolve-then-construct approach inevitably runs into the reality gap \cite{jakobi1995noise}. On the other hand, there exist systems where physical objects are evolved in the real-world, but these are passive artefacts without agency \cite{Rieffel-Sayles-2010,Kuehn-Rieffel-2012}. Complete systems where real robots undergo evolution are still ahead of us, but with the development of 3D-printing, rapid prototyping, and automated assembly such systems are becoming feasible, at least in an academic setting \cite{brodbeck2015morphological,hale2019robot, jelisavcic2017real,Vujovic2017}. 

\begin{figure}[!htbp]
    \centering
    \includegraphics[width=.33\textwidth]{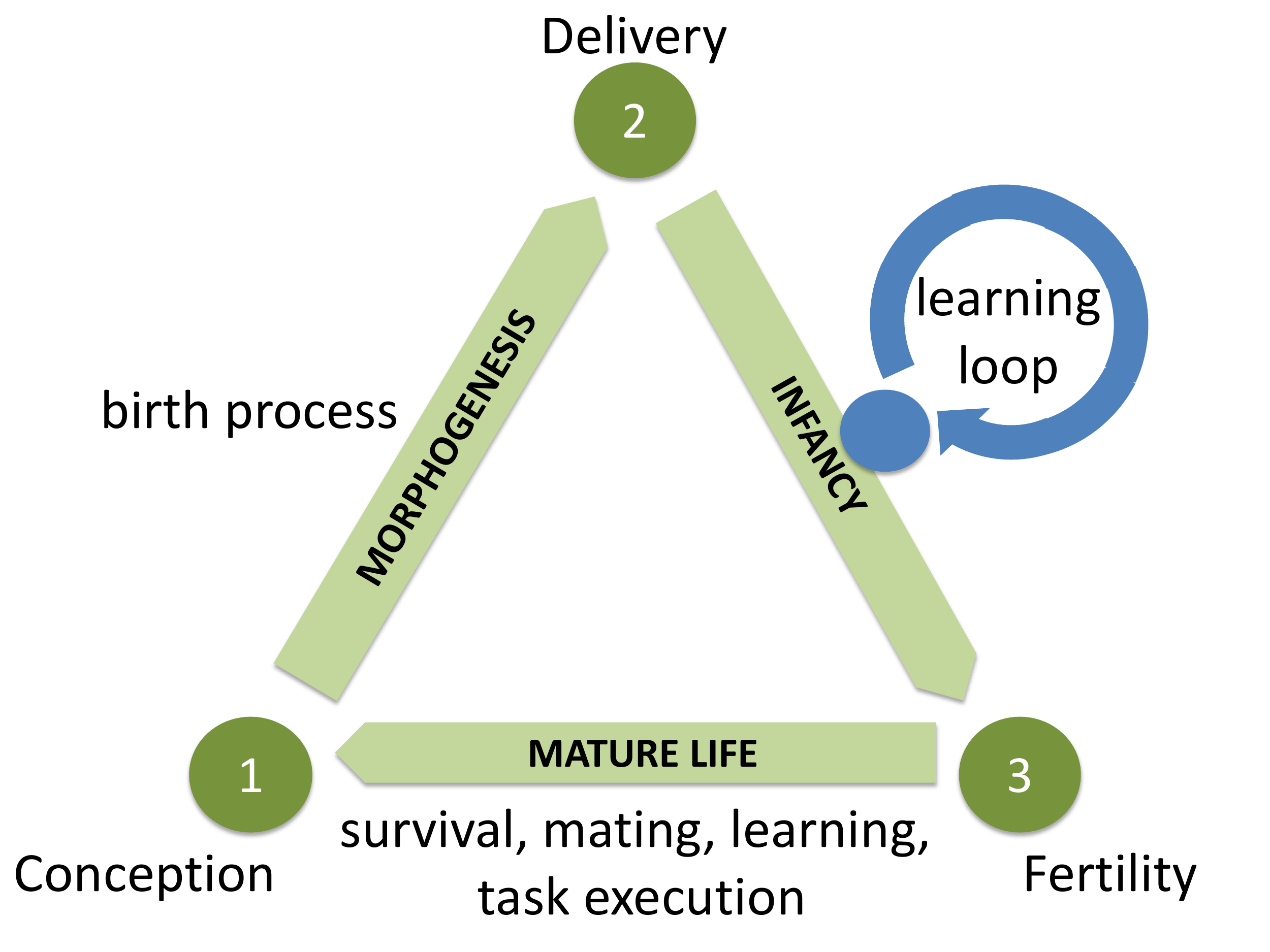}
    \caption{Generic system architecture for robot evolution conceptualized by the \tol{}.
    }
	\label{fig:system}
\end{figure} 

A generic model to underpin ``evolving robots in real time and real space'' has been described recently \cite{eiben2013triangle}. 
This model, called the \tol{}, captures the universal life cycle of a robot, not from birth to death because that would not be a cycle, but from conception (being conceived) to conception (conceiving offspring). This cycle consists of three stages: morphogenesis (from conception to birth), infancy (from birth to becoming a fertile adult), and mature life (during which the robot can mate and conceive offspring multiple times), cf. Figure \ref{fig:system}. 


A key insight behind this paper is that including a learning stage is not an arbitrary design choice, it is pivotal for mitigating a general problem. Namely, while it can be assumed that the parents had well-matching bodies and brains (otherwise they had not been fit enough to be selected for mating), in general it cannot be assumed that crossover preserves the good match. The mis-match in the offspring may be severe (e.g., having more sensors than there are inputs in the controller) or moderate (e.g., only requiring some parameter tuning in the brain), but in any case it is important to adjust and optimize the inherited brain quickly after `birth'. 

The problem we highlight here is inherent to morphological robot evolution where a large number and variety of robots is produced through consecutive generations. All these robots with different and unpredictable morphologies have to learn the tasks required by the given application. Thus, a morphologically evolving robot system needs a learning method that works in any of the possible robots and finds a good controller efficiently. 

Comparing with current studies, the main contributions of the paper are the following. 
Firstly, a new type of controller architecture based on coupled oscillators with sensory feedback that can be customized to any modular robot driven by joints. 
In such a type of controller, a generic method is used to specify a frame of reference (coordinates for the robot modules) for any given body in our design space.
A steering mechanism based on scaling the activation signals, depending on the coordinates of the joint and the angle between the direction to the target and the robot's heading, is used to drive the robot joints. 
These controllers can be used in a closed-loop approach and steer a robot towards a target regardless of the specific shape of modular robots. 
This makes it possible to generate a closed-loop controller for a modular robot with an arbitrary shape.
Secondly, a generic learning method that allows modular robots with arbitrary shapes to learn approaching a target. 
We validate our method with three different robots, rather than a fixed and special shape robot, in three different scenarios. To this end, we use three modular robots, a `spider', a `gecko', and their `baby' created by an evolutionary robotics project in our lab \cite{jelisavcic2017real}.

\begin{figure}[!ht]
    \centering
    \includegraphics[width=0.49\textwidth]{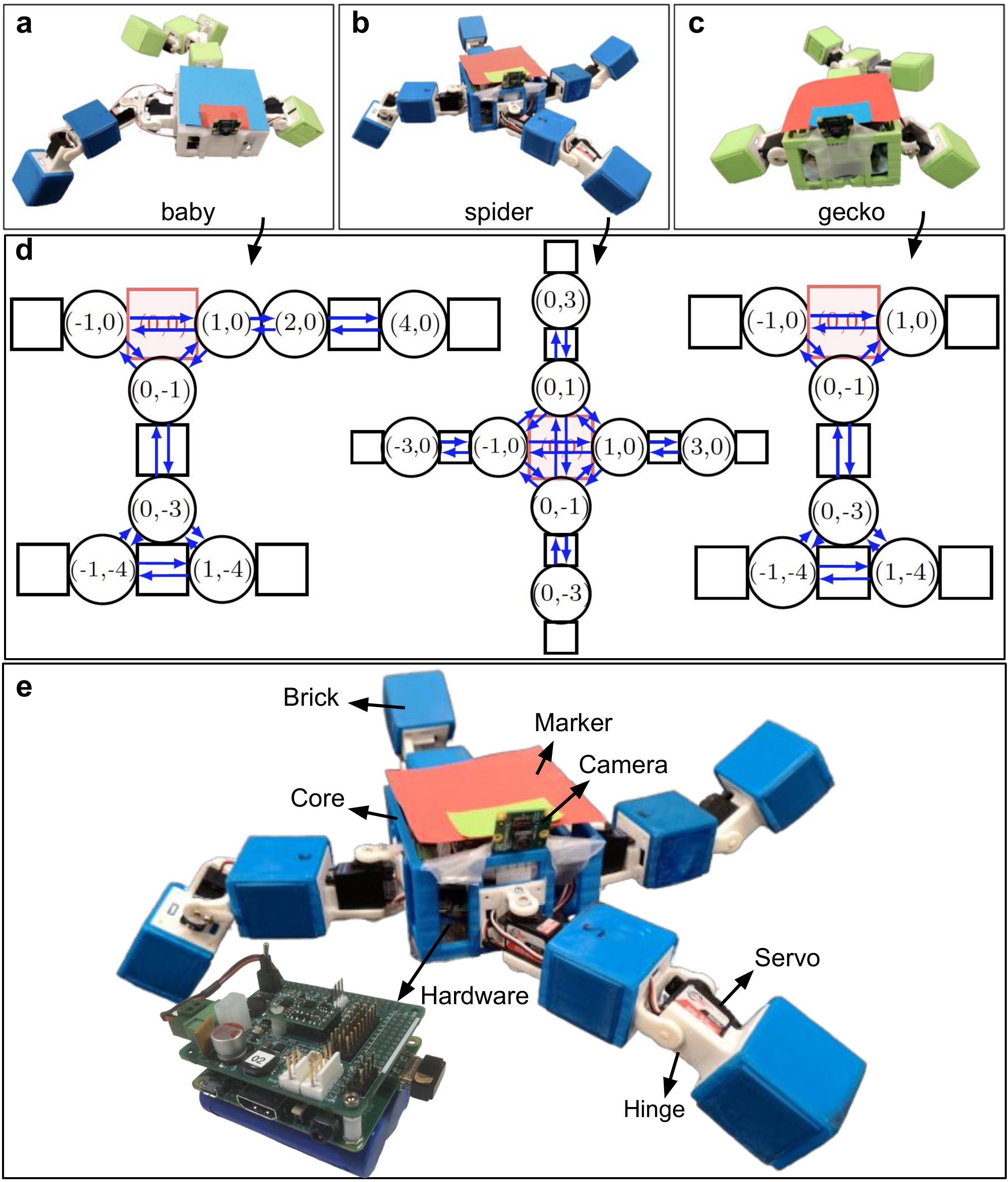}
    \caption{The three robots in our test suite after \cite{jelisavcic2017real}. The spider (b) and gecko (c) are the parents, the baby (a) is the offspring. (d) exhibits the overall network topologies of the controllers and the coordinates of the joints in the baby (left), spider (middle), and gecko (right) robot. The core block (head) is red, other blocks are black. Joints are represented by circles.
    (e) shows the inner components of the spider robot.}
    \label{fig:robots}
\end{figure}



\section{Good locomotion and how to learn it}
\label{sec:loco-learn}

The existing studies in learning locomotion on modular roobts can be divided into two categories in terms of the controllers, open-loop and closed-loop. The dominant approaches are using Central Pattern Generators (CPGs) \cite{Ijspeert2008learning}.
Most papers studied gait learning with real robots using an open-loop controller with no sensory feedback from the environment \cite{Kamimura2005Automatic,Bongard2006Science,Kyrre2015real-world,Thakker2014ReBis}. 
For closed-loop controllers, joint angle and foot contact are typically used to be the sensory feedback. 
The studies \cite{owaki2013simple,owaki2017quadruped,nordmoen2019evolved} used a CPG-based controller and foot contact from force sensors on each robot leg to produce coordinated gaits and increase its adaptability on various terrains. 
Inertial measurement \cite{wang2005motion,seo2010cpg,barasuol2013reactive,sartoretti2018central}, joint angles \cite{kimura2007adaptive}, and touch sensing \cite{righetti2008pattern,ajallooeian2013central} are used to be sensory feedback in CPG-based controllers to adjust robot behaviours for desired tasks. 
In particular, the combination of a CPG-based controller and camera is used to achieve the closed-loop control for directed locomotion in a hexapod robot \cite{shaw2019workspace}.
Similarly, there are other studies \cite{wu2013neurally,Ijspeert2007From} that use various sensory feedback and controllers to achieve closed-loop control for different tasks.
Although these studies proposed closed-loop controllers with various sensory feedback for locomotion, they focus on learning locomotion on a fixed shape robot such as a hexapod robot, a fish robot, a salamander robot. It is still generally lacks generic methodologies for integrating sensory feedback to adapt the locomotion for a modular robot with an arbitrary shape.

In this paper, we consider the problem of targeted locomotion on modular robots with arbitrary shapes. This is important, challenging, and novel. Targeted locomotion is important, because for many applications robots should be capable of going to a point of interest, be it a charging station, an object to fetch, or another robot to mate with. Targeted locomotion on evolvable modular robots is challenging because the number and the spatial arrangement of the joints, the length and branches of the limbs can vary and the overall shape can be irregular. This makes the simple adoption of the steering policy for wheeled robots --to turn left (right) apply more force to the wheels on the right (left)-- highly non-trivial. Finally, this problem is also novel; to our best knowledge there are no publications addressing this. 


\subsection{Robots with a sense of direction}
\label{sec:loco}

To enable targeted locomotion of modular robots with arbitrary morphologies, we need a generic way of defining a \textit{frame of reference}. To this end we define a coordinate system based on the fact that our robots have a core module with a camera, see Figure \ref{fig:robots}. By definition, the `head', that is, the core module is the origin $(0,0)$ and the direction of the camera determines `North'. Now we can assign the coordinates $\{(1,0), (2,0), ...\}$ and $\{(-1,0), (-2,0), ...\}$ to the modules (passive bricks or active joints) as we go to East and West, respectively.
Similarly, if we go North and South we set the coordinates to $\{(0,1), (0,2), ...\}$ and $\{(0,-1), (0,-2), ...\}$, respectively. The coordinates of the three robots in this paper are shown in Figure \ref{fig:robots}.



To make robots see the target we use a system proposed in \cite{lan2018ICARCV}. When a target is detected its direction $\alpha \in [-\beta, \beta]$ w.r.t. the robot can be determined, where $[-\beta, \beta]$ is the cameras field of vision. This information can be combined with the frame of reference defined above: If $\alpha < 0$, then the target is on the left; otherwise it is on the right.

Following the literature we use coupled oscillators to control the joints of a robot and we arrange these oscillators in a network to form Central Pattern Generators (CPGs) \cite{ijspeert2008central, steuer2019central}. Such oscillators generate tightly-coupled patterns of neural activity that drive rhythmic and stereotyped locomotion behaviors like walking, swimming, flying in vertebrate species and they have been proven to perform well in modular robots as well \cite{hultborn2007spinal,ijspeert2008central,Ijspeert2007From}.

\begin{figure}[!htbp]
	    \centering
	    \includegraphics[width=\columnwidth]{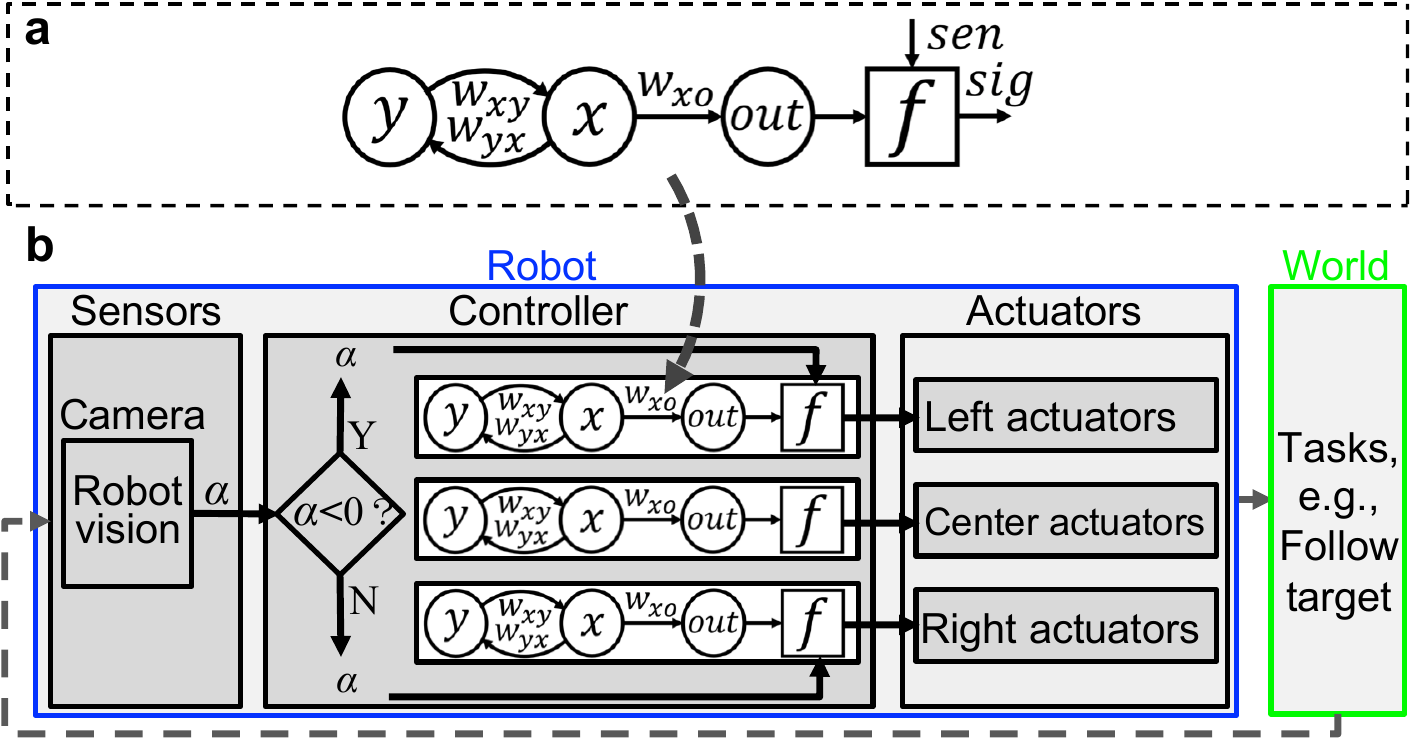}
		\caption{(a) A sensory oscillator with three neurons, $x$-neuron, $y$-neuron, $out$-neuron and an extra node $f$. The function $f$ combines the raw control signal of the $x$-neuron with the sensory information $sen$. (b) Overall scheme of steering by sending control commands to actuators depending on their lateral position and the direction of the target.}
		\label{fig:framework}
\end{figure}

To provide useful information from the environment we introduce the notion of a sensory oscillator as shown in Figure \ref{fig:framework} (a). 
Such a sensory oscillator drives one joint in the robot by two coupled neurons $x$ and $y$, and an $out$-neuron that together generate a signal regulated by three weights as usual. 
Subsequently, the extra (square shaped) node generates the actual signal $sig$ for the joint by applying a function $f$ to the sensory information $sen$ and the signal of the $out$-neuron.
The novelty of this model lies in the extra node that considers the sensory input before generating the signal that actually drives the joint. The model is general, there is no restriction on the sensory input(s) and $f$ can be any function depending on the task at hand.

The subsystems described above can be combined into an adequate control system that enables the modular robots to move towards a target. To this end, we use the angle to the target $\alpha$ as input and generate a scaling factor $d_p(\alpha)$ and define a function $f$ to adjust the signals of the joints on the left and right side of the robots as necessary. The exact details are described in the Methods section, the overall effect is that if the target is on the left (right), then the joints on the left (right) apply less force and make the robot turn in the correct direction. The middle joints are never modified. The overall architecture is exhibited in Figure \ref{fig:framework} (b).
The overall control system of a robot is a CPG network where sensory oscillators of neighboring joints are connected as shown in Figure \ref{fig:robots} (d).

\subsection{Learning method}
\label{sec:learn}

Learning a task in our system amounts to finding proper weights for the controller. This boils down to optimizing a black-box objective function. Bayesian optimization (BO) is the state-of-the-art method in terms of data efficiency, but its computational complexity grows cubically with respect to the number of observations \cite{jasper2012practical}. 
Alternatively, Evolutionary Algorithms (EA) take constant time for generating candidate solutions \cite{eiben2015evolutionary}. This makes their overhead much less computation-intensive than that of the BO, at the expense of data efficiency. 

To get the best of both worlds, here we use a combined algorithm, the Bayesian Evolutionary Algorithm (BEA), that starts with BO and runs this until the time efficiency becomes too low. At this point, a certain subset $S$ of the solutions generated so far is selected by considering their quality and diversity. This set $S$ is then used as the initial population for the EA that is run until a good solution is found or the given computing budget is exhausted. We refer the reader to the  Methods section for further details.

%

\section{Experiments}
\label{sec:experiments}
We employed the BEA to learn adequate controllers in simulation. Then we tested the learned controllers in three test scenarios: 1) approaching a fixed target, 2) following a moving target, and 3) following another robot that is following a moving target. Let us note that the robots do not change or tune their controllers between scenarios. 

We used the customized framework \emph{Revolve} \cite{hupkes_2018_revolve} to learn appropriate controllers for directed locomotion at a zero angle (thus: straight forward) on the virtual gecko, spider, and baby robots.
The fitness function for the BEA was a combination of the deviation from the required direction and the distance covered during the test period, see the Methods section for details. The duration of a test period was 60 seconds and the BEA was allowed to perform 1500 evaluations. 
Despite the heavy simulations, the computing times were acceptable, approximately one hour was enough to complete one run on a Linux PC with a 3.0GHz CPU, 64Gb RAM, and 32 cores with multi-threading. Conducting the whole learning process on the real robots would take several days, $1500 \times 60$ seconds, plus overhead for re-positioning the robots between tests, charging the batteries, and handling breakdowns.

\subsection{Scenario 1: fixed target}
\label{sec:fixed}
\begin{figure}[!htbp]
	\centering
	    \begin{adjustbox}{max width=0.49\textwidth}
	\begin{tabular}{c c c}		\includegraphics[width=0.4\textwidth]{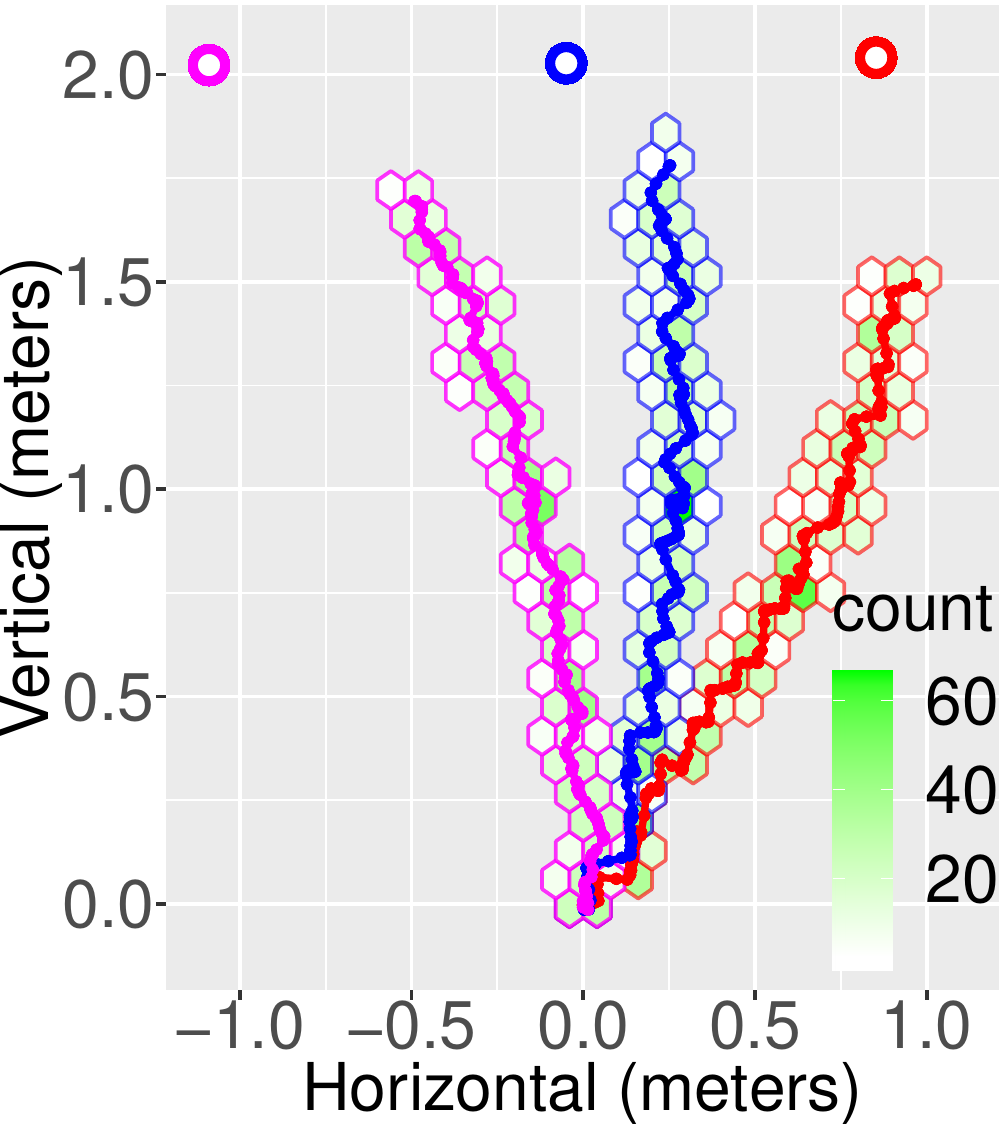}&
		\includegraphics[width=0.4\textwidth]{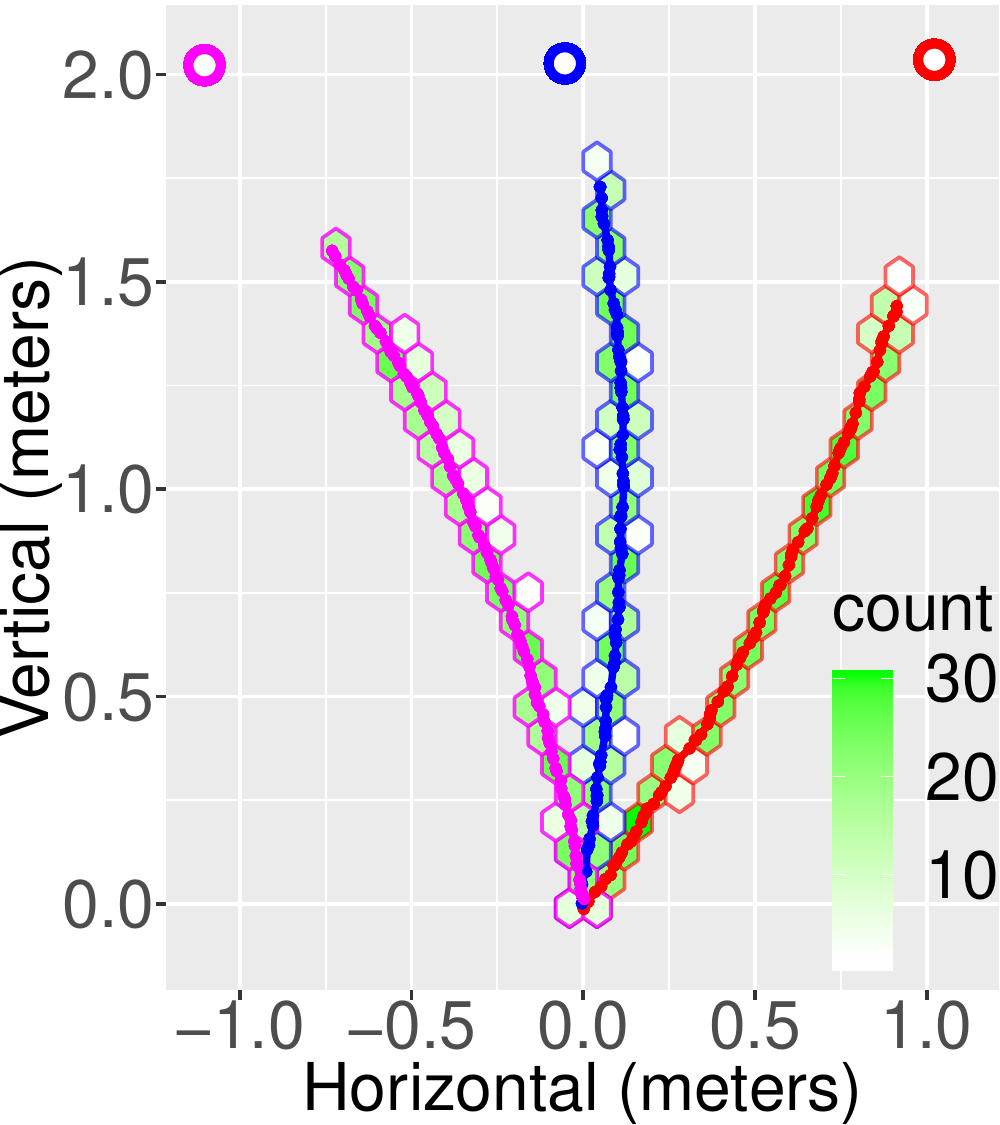}&
		\includegraphics[width=0.4\textwidth]{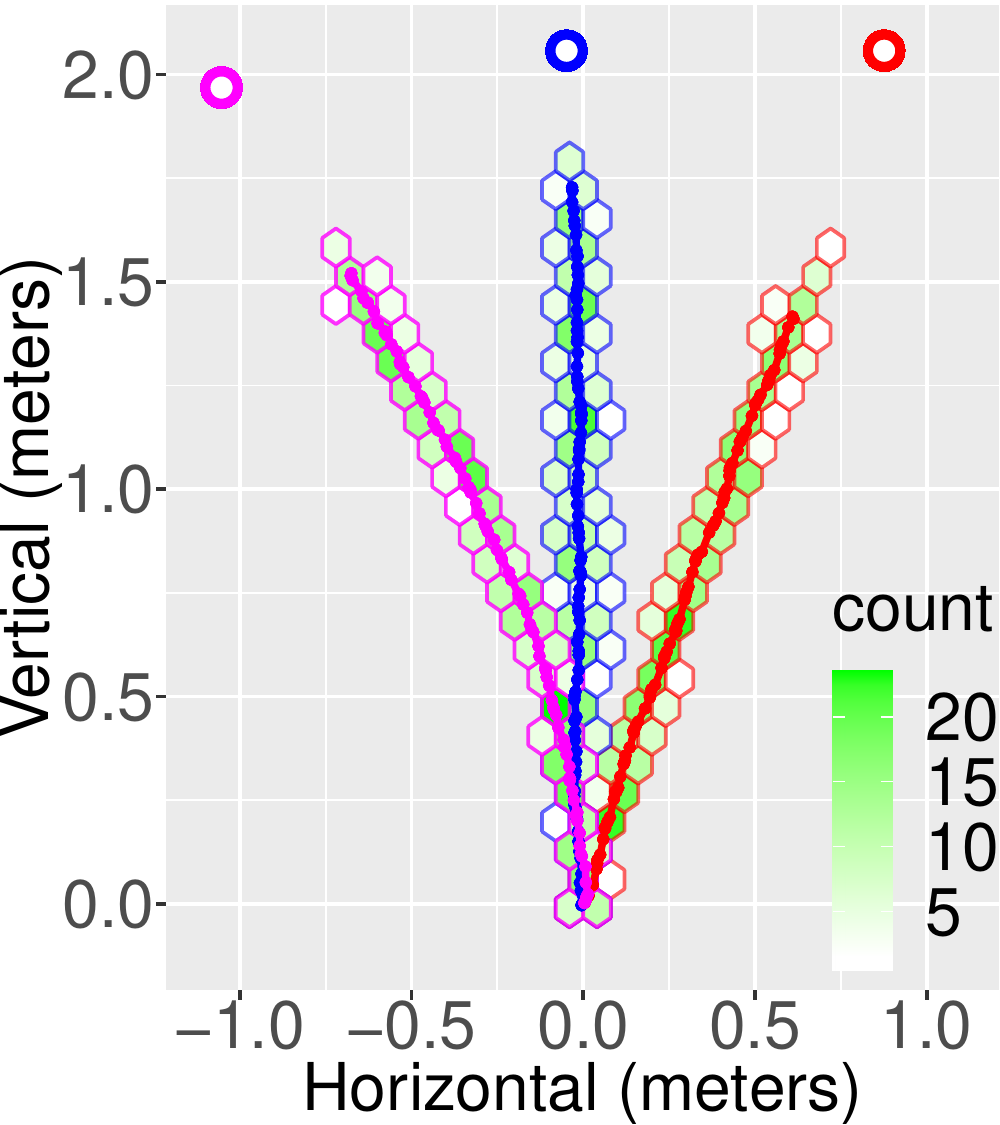} \\
	\end{tabular}
    \end{adjustbox}
    \begin{adjustbox}{max width=0.49\textwidth}
	\begin{tabular}{c c c}
		\includegraphics[width=0.4\textwidth]{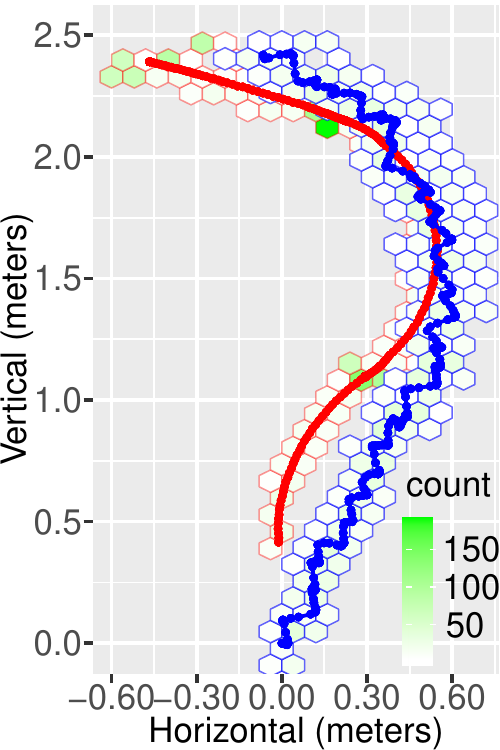}&
		\includegraphics[width=0.4\textwidth]{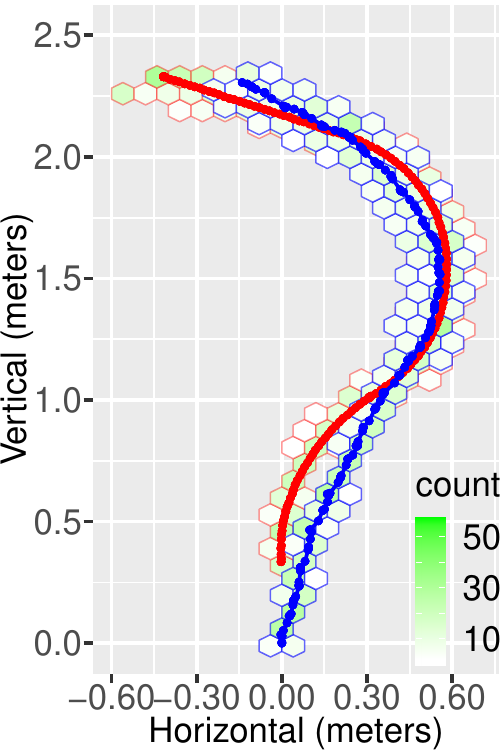}&
		\includegraphics[width=0.4\textwidth]{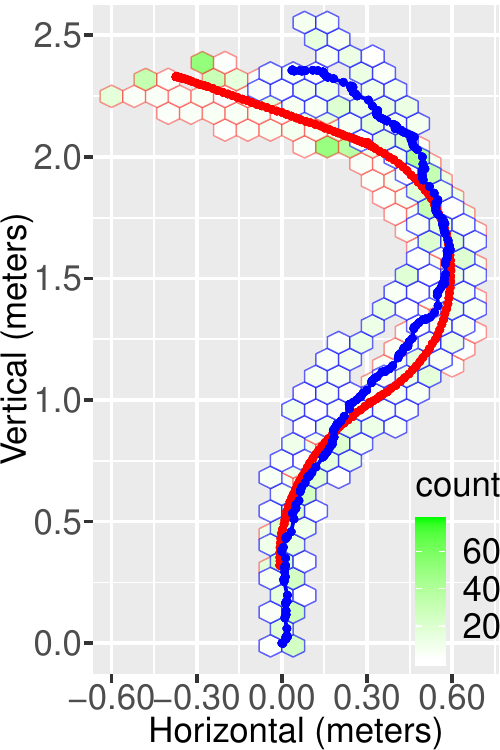} \\
	\end{tabular}
    \end{adjustbox}
	\caption{Trajectories of the spider (left), the gecko (middle), and the baby (right). The solid lines show the average trajectories (five runs). The hexagonal bins show the number that robots located in the hexagonal area for all five repetitions, where the location of robots are collected per 0.5 second. Top row: moving towards a fixed target, the coloured circles. Bottom row: following a moving target. The red line shows the path of the target robot, the blue one belongs to the `chaser'.  
	}
	\label{fig:trajectories}
\end{figure}

The controllers learned in simulation are validated on the real robots, the spider, the gecko, and the baby. In the first series of experiments, we tested each robot with three different setups, one with a fixed target to the left, one with the target straight ahead of the robot, one with the target to the right. In all cases, the target was in the field of view of the robot camera (Raspberry Pi Camera Module v2) at the start of the test. We repeated each test five times and displayed the observed trajectories in Figure \ref{fig:trajectories}.
The experiments were recorded with an overhead camera above the test arena, as shown in the supplementary videos.
The plots show a consistent behavior for all robots and test cases. The path of the spider shows that it is `wobbling', see the top left plot in Figure \ref{fig:trajectories}. This is understandable because its middle-line lies over two of its limbs, hence straight forward movement is very hard without zigzagging. Comparing the blue curves with the other colors we can also see that the robots approach the target in the center faster. This result can be easily explained, because for a target on one of the sides, the robot must turn and that costs extra time.

\subsection{Scenario 2: moving target}
\label{sec:moving}
In the second series of experiments, we tested each robot with a moving target. To this end, we used a wheeled Robobo robot that was pre-programmed to drive a given trajectory. In the initial position, the Robobo was approximately 30 cm ahead of the modular robot and started to drive to the right, then it turned and drove to the left. The bottom row of Figure \ref{fig:trajectories} shows the trajectories after five repetitions with each modular robot. Additionally, we recorded the experiments with an overhead camera above the test arena (see Supplementary Video). These data indicate that the modular robots were able to follow the Robobo in all cases. This demonstrates that the controllers learned for a simple task (moving straight ahead) were applicable in a different and more difficult task. In turn, this proves the usefulness of our new controllers based on an internal frame of reference and sensory oscillators. 


%
\begin{figure*}[!ht]
	\centering
    \begin{adjustbox}{max width=0.95\textwidth}
	\begin{tabular}{c c c c c}
		\includegraphics[width=0.385\textwidth,angle=90]{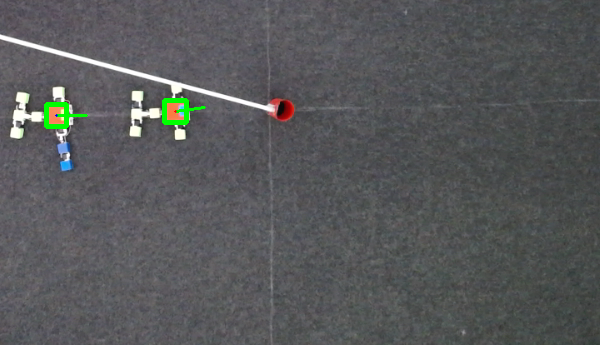}& \hspace{-4mm}
        \includegraphics[width=0.385\textwidth,angle=90]{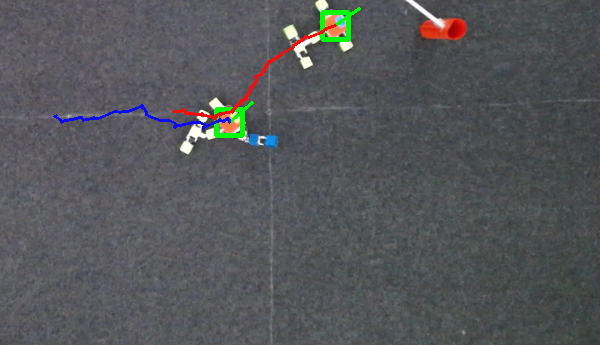}& \hspace{-4mm}
		\includegraphics[width=0.385\textwidth,angle=90]{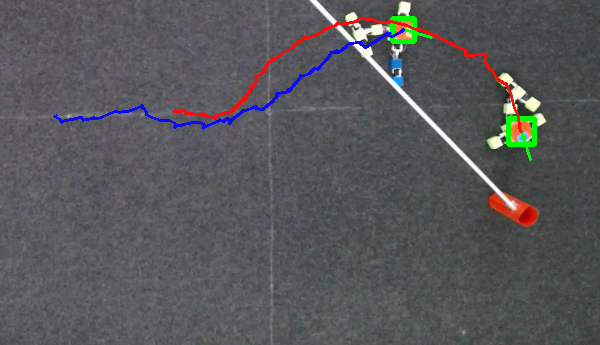}& \hspace{-4mm}
		\includegraphics[width=0.385\textwidth,angle=90]{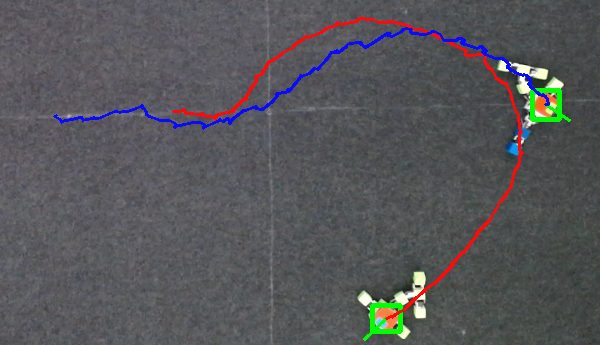}& \hspace{-4mm}
		\includegraphics[width=0.385\textwidth,angle=90]{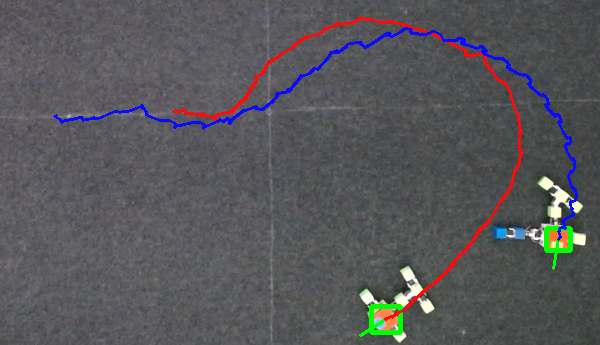} \\
		\includegraphics[width=0.363\textwidth,angle=90]{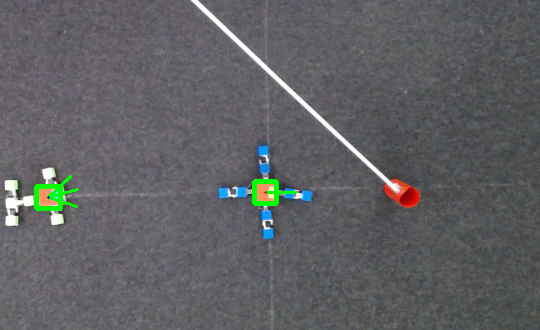}& \hspace{-4mm}
		\includegraphics[width=0.363\textwidth,angle=90]{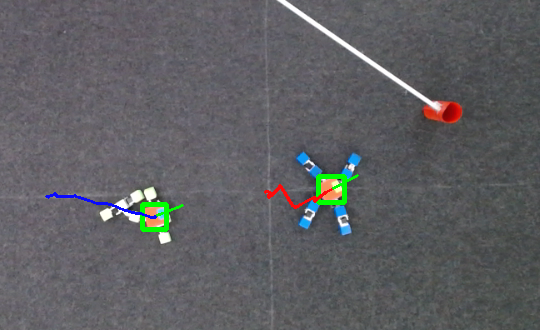}& \hspace{-4mm}
		\includegraphics[width=0.363\textwidth,angle=90]{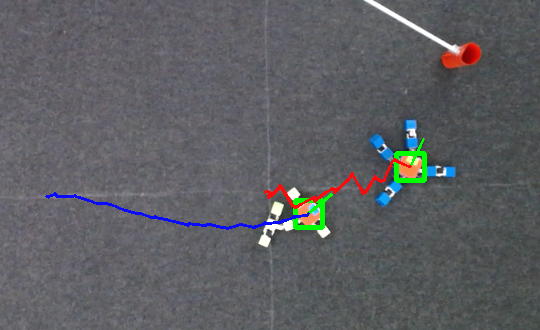}& \hspace{-4mm}
		\includegraphics[width=0.363\textwidth,angle=90]{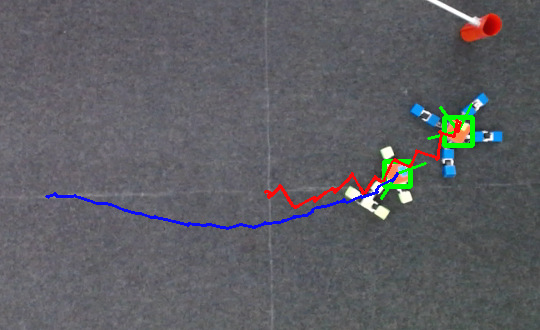}& \hspace{-4mm}
		\includegraphics[width=0.363\textwidth,angle=90]{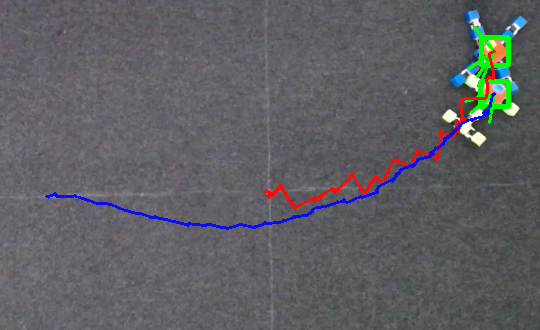} \\
	\end{tabular}
    \end{adjustbox}
	\caption{Still images of the double moving target experiments. Top row: the gecko follows the hand-held target and the baby follows the gecko. Bottom row: the spider follows the hand-held target and the gecko follows the spider. The hand-held target is the red cylinder at the end of a white stick. The red and blue lines show the trajectories of the first and the second robot, respectively.}
	\label{fig:double-overhead}
\end{figure*}

\subsection{Scenario 3: double moving target}
\label{sec:d-moving}

In the third series of real-world experiments, we challenged the modular robots even further. In this setup, we replaced the Robobo used in the previous experiments by one of the modular robots and hand-held a target in front of it. Another modular robot was placed on the regular starting position. Then the robots started at the same time, the first one following the target hand-held by the experimenter, the second one following the first one. In Figure \ref{fig:double-overhead}, we show two of these tests, the case of the gecko following the hand-held target and the baby following the gecko and the case of the spider following the hand-held target and the gecko following the spider. 
The figure captures the test by five still images taken by the overhead camera, more information is shown in Supplementary Video. These show that robots can follow a target even if it is an irregularly moving irregular shape (another modular robot). The trajectories also disclose the speed differences. In the first test, the distance between the gecko and the baby is gradually growing, which indicates that the baby is a little slower than the gecko. In particular, the baby encountered a little difficulty in turning right. This is not surprising, given that its morphology with the long right limb is not perfectly symmetrical. In the second test we see the opposite, the second robot is closing in on the first one, and before the end of the experiment the gecko hits the spider.


\section{Discussion}
\label{sec:discussion}


The wider context of this study is full-fledged robot evolution, where both morphologies and controllers evolve. We note that a morphologically evolving robot system needs a learning component that works in any of the possible robots and adjusts the inherited controller (or a randomly initialized controller) to the given morphology and task. 

Our experimental work is, in essence, a feasibility study addressing this general issue within our system of evolvable robots and a specific task. 
The concrete problem we tackle is to enable targeted locomotion, that is, approaching and following an object, in modular robots without any assumption on their specific morphology. This forms a good test case because it is morphology dependent, challenging for robots with random shapes, and practically relevant. The solution we deliver is based on an internal frame of reference and the use of oscillators with sensory feedback to drive the joints. These are combined into a control system that modulates the force in the servo motors depending on the robots angle to the target and the location of the joint in the robots body. The overall control system is a CPG-based network where neighboring joints of the robot are connected. The other ingredient of the solution is a learning method that can find good parameter values for any given CPG-based network. 

To validate our approach we designed nine test cases by three robots with different morphologies and three scenarios, one with a fixed target, one with a moving target, and one with a double moving target. The robots learned a good controller for moving straight forward and these controllers were evaluated in the test scenarios.

In the first scenario, all robots approached the target accurately, even though the spider exhibited a little offset to the right, cf. Figure \ref{fig:trajectories} and Supplementary Video. In the second scenario, the target moved to the right first and turned to the left halfway the test. All three robots could adjust their trajectory and followed the target with a little delay as can be seen in Figure \ref{fig:trajectories}. This proves that the sensory-motor feedback loop they learned for one basic skill (straight forward movement) was generalizable to a more dynamic and complex task.

In the third scenario, a modular robot had to follow another modular robot that followed a hand-held target. This test further confirmed the adequacy of the learned controllers and illuminated the differences between the locomotion abilities. For instance, the baby followed the gecko well in a left turn, but it fell behind when turning right because of the longer right-front limb as shown in the fifth frame of Figure \ref{fig:double-overhead}.
In the other test, the gecko followed the spider easily because of the wiggly locomotion of the spider and the relatively high speed of the gecko.

The results of the experiments demonstrate that the new type of oscillators and CPGs with external feedback in combination with the internal frame of reference empower robots with different morphologies to perform object following. Our frame of reference is in essence a 2-dimensional coordinate system, where the origin and the `North' are defined by specific features in our robots, the head module and the direction of the camera, respectively. This concept can be easily extended to three dimensions and to different robots with other features. For instance, a designated head module is not required, the origin can be defined by any reasonable principle that applies to the given morphologies. Likewise, while using the direction of the camera to define `North' is a natural choice in our application, a coordinate system can be defined for robots without a camera as well. 

Further extensions and generalizations are possible regarding the feedback from the environment. Although in this study we use a camera, our approach is applicable in a wide range of robot systems, because the sensory oscillators can handle inputs from different sensors, e.g., accelerometers, gyroscopes, IR sensors, sonars. Furthermore, in this paper all joints on the same side are treated the same way, cf. Equations \ref{eq:l_joint_downscale} and \ref{eq:r_joint_downscale}, but it is possible to define a variant where the exact coordinates, for instance the distance to the `spinal cord', are also taken into account. 

Last but not least, improvements are possible by employing another learning method. The BEA we use here is not application specific, in principle any derivative-free black-box optimization algorithm for learning the adequate parameter values for the controller is applicable. In our view, it is very important to consider both sample efficiency and time efficiency, that is, the number of trials or evaluations as well as the time needed to achieve a decent result. In this paper we used a budget of 1500 trials and the BEA spent about one hour on learning. This is certainly acceptable and delivers the proof of concept we aimed for, but we are convinced that these figures can be improved by more advanced methods. 

A particular aspect here is the dichotomy between the real-world and simulations. To this end, it is important to note that the use of simulations does not invalidate the concept of physical robot evolution. In fact, simulations can be used in both the evolutionary and the learning loop.
As outlined in \cite{howard2019evolving} and \cite{hale2019robot} there are great potential advantages in evolving real and simulated robots simultaneously. If the simulated and real robots share the same genetic language then we can `cross-breed' them, using a virtual `mother' and a physical `father'. In such a hybrid evolutionary system physical evolution is accelerated by the virtual component that can find good robot features with less time and resources than physical evolution, while simulated evolution benefits from the influx of genes that are tested favourably in the real world. Additionally, even if we have clean physical evolution and all robot `children' are produced in the real world, the learning process in the Infancy stage can use simulations to obtain a good controller for the given robot `child'. The learned controllers can suffer from the reality gap, which can be mitigated by a subsequent real-world learning process on the physical robot. The advantage of the combination is that virtual learning can be on a meso-scale, spending quite a few trials (e.g., hundreds, like in our case) and the real-world process on a micro-scale with much fewer trials (perhaps just a few dozens). For the sake of the argument, if 100 real-world trials were enough, then the learning process could be completed on the real robots in an afternoon. The current paper is on the simulated side. The controllers learned in simulation just worked for us here, thus we had no need to add a real-world learning process.


Reflecting from a broader perspective, this work provides an approach for learning tasks inside an evolutionary process that produces various morphologies. Our approach is generic, applicable to various types of evolvable robot systems and different tasks. For example, it can be applied in a system for evolving robots for exploring and decommissioning nuclear power plants \cite{hale2019robot}, as well as in more fundamental studies regarding the evolution of embodied intelligence. 






\section{Methods}
\label{sec:methods}

\subsection*{The general system}
\label{subsec:general_framework}

The robots in our system are based on RoboGen \cite{auerbach2014robogen}, they consist of a core component that hosts the controller board, the battery and a camera, 3D-printed passive bricks, and joints driven by servo motors \cite{jelisavcic2017real}. Figure \ref{fig:robots} displays the three robots we use in the current experiments. The camera provides information about the environment that is processed by a new type of control system based on sensory oscillators that activate the servos. 
A schematic representation of the system is presented in Figure \ref{fig:framework} (b). 
In the next subsections, we discuss the details of each component of the system.
%
%
\subsection*{Robot Vision}
\label{subsec:robot_vision}

The closed-loop controllers are based on visual input delivered by a camera. The robot vision system must work accurately in real-time on our Raspberry Pi Camera Module v2, and ideally be power efficient. The two pivotal steps for this system are the recognition of objects of interest, and the calculation of the angle of a target object w.r.t. the orientation of the robot.

\paragraph{Object recognition} 
Here, we follow the approach proposed in \cite{lan2018ICARCV,lan2019evolving} that allows to quickly recognize targets with low-performance hardware.
The method consists of two components:
\begin{itemize}[nolistsep]
    \item Detection of region of interests (ROIs) using \textit{Fast ROIs Search} proposed in \cite{lan2018ICARCV}.
    \item Object recognition using Histogram of Gradients (HOG) as a feature extractor and Support Vector Machines (SVM) with the linear kernel as a detection method.
\end{itemize}
We decided to use the combination of HOG and SVM instead of Deep Neural Networks since HOG and SVM perform much faster. In this project, we used the implementation provided by OpenCV.

\paragraph{Angle to target object}
Once a target is detected, its relative position from the robot's perspective can be expressed by the angle $\alpha \in [-\beta, \beta]$, where $\beta$ is the angle determined by the field of vision of the given camera. The angle to the target $\alpha$ is the angle between the orientation of the robot (i.e, the camera) and the target object. If the target is on the left-hand side of the robot’s face, then $\alpha$ is negative, whereas if the target is on the right-hand side of the robot’s face, $\alpha$ is positive. The value of $\alpha$ is zero if the target is straight ahead of the robot. 
Given an image registered by the Raspberry Pi Camera Module v2 with the parameters (the field of view ($2 \times \beta$) is $\ang{62.2}$, the resolution is $3280 \times 2464$ where 3280 is the number of pixel columns ($\mathcal{N}_c$)), where the coordinate of the target can be recognized by the robot vision as $(x,y)$ in pixels, $\alpha$ is calculated from this image by
\begin{equation}
    \alpha = \frac{\arctan(\frac{x - \mathcal{N}_c / 2} {\mathcal{F}})} { \pi \times 180 }
\end{equation}
where $\mathcal{F}$ is a potential inherent factor that depends on the camera, can be calculated by 
\begin{equation}
    \mathcal{F} = \frac{ \mathcal{N}_c /2}{\tan(\frac{\beta}{180 \times \pi})}
\end{equation}

Due to the limited camera's field of view in the real world, the robots have to process the situation that the target is out of the camera's field of view. 
For such a situation, we expect the robots to search the target until the target is in the camera's field of view.
To this end, we use two solutions: 1) If the target is out of the camera's field of view initially, the robots have to search the target until it is in the camera's field of view. For this purpose, the initial value of $\alpha$ can be set as $\beta/2$ (or $-\beta/2$) for turning right (or left).
2) If the target escapes from the camera's field of view during locomotion, the robots keep the previous behaviours to follow the target until the target comes back to the field of view. 
That is, $\alpha$ preserves the previous value if no target is in the field of view except the initial stage.
\subsection*{Controller}
\label{subsec:controller}

\paragraph{Modeling joints as oscillators}
The key element of the controller is the model of a single joint. Here, we propose a new oscillator with sensory feedback to properly represent oscillatory behavior often seen in nature \cite{ijspeert2008central}.
The controller composed of new oscillators works in a closed-loop, in which the robot's action can be changed according to the sensory feedback about the target in the environment.
In general, the sensory input to the new oscillator can be generated by any sensors.

A sensory oscillator has an $x$-neuron, a $y$-neuron, an $out$-neuron, and an extra node that implements a function $f$.
For each time step, neuron $x$ ($y$) feeds its activation value multiplied by the weight $w_{xy}$ ($w_{yx}$) to the neuron $y$ ($x$).
At a time step $t$, the changes of the activation value of an $x$-neuron and a $y$-neuron can be calculated as $\Delta x_{(t)}  = w_{yx}y_{(t-1)}$ and $\Delta y_{(t)}  = w_{xy}x_{(t-1)}$ respectively, 
where $t-1$ represents the last time step. 
The $x$-neuron and the $y$-neuron generate the activation values $x_{(t)}$ and $y_{(t)}$ of oscillatory patterns over time according to the following expression:
\begin{align}
    \begin{split}
        x_{(t)} = x_{(t-1)} + \Delta x_{(t)} \\
    	y_{(t)} = y_{(t-1)} + \Delta y_{(t)}  
    \end{split}
    \label{eq:activation}
\end{align}
%

\par The $x$-neuron feeds its activation value multiplied by the weight $w_{xo}$ to the $out$-neuron.
The $out$-neuron applies an activation function to generate the activation value.
For the oscillator in the CPG-based controller of modular robots with joints driven by servo motors, the activation values of the $out$-neurons have to meet two conditions due to the limited rotating angle of the joints. 
First, the activation value of $out$-neuron must be periodic that repeatedly returns to its initial condition.
According to the stability criterion for linear dynamical systems \cite{bubnicki2005modern}, it is beneficial to take $w_{yx} = - w_{xy}$. Such values of parameters lead to periodic signals that do not explode over time.
In this study, we use the predefined initial values $(x_{(0)}, y_{(0)}) = (-\frac{1}{2}\sqrt{2}, \frac{1}{2}\sqrt{2})$, and $(w_{xy}, w_{yx}) = (0.5, -0.5)$, but they can be randomly initialized except 0. 
Second, the activation value of $out$-neuron should be bounded in an interval.
Therefore, we use a variant of the sigmoid function, the hyperbolic tangent function ($tanh$), as the activation function of $out$-neurons to bound the output value in $[-1,1]$.
At a time step $t$, the $tanh$ activation value of $out$-neuron can be calculated as follows:
\begin{equation}
    out_{(t)} = \frac{e^{x_{(t)}} - e^{- x_{(t)}}}{e^{x_{(t)}} + e^{- x_{(t)}}} .
    \label{eq:output}
\end{equation}
%


Finally, the new oscillator executes an extra operation $f$ that combines the activation value of $out$-neuron and the external sensory signal $sen$ to produce a signal $sig$, i.e., $sig = f(sen, out)$.
In general, $f$ could be any function, here we use multiplication of $out$ and $sen$.
Hence, at a time step $t$, the $sig$ value can be calculated by:
\begin{equation}
    sig_{(t)} = sen_{(t)} \times out_{(t)}
    \label{eq:multiplication}
\end{equation}
The rationale behind our new oscillator model is to allow the inclusion of a sensory signal for a closed-loop control. We refer to this new model as the sensory oscillator (see Figure \ref{fig:framework} (a)). 


\paragraph{Network of oscillators}

For modular robots, e.g., the robots in Figure \ref{fig:robots}, there are multiple joints affecting each other for achieving the actions.
In other words, we deal with a network of joints (oscillators) and we must take into account the influence of other oscillators.
In the CPG-based network controller, the neighboring oscillators are connected to each other as the blue arrow lines shown in Figure \ref{fig:robots} (d).
As a result, we have a further expression for a single oscillator by including the neighborhood instead of Equation \ref{eq:activation}.
For the $i$-th sensory oscillator, the activation values $x_{i(t)}$ of its $x$-neuron and $y_{i(t)}$ of its $y$-neuron can be calculated as:
\begin{align}
    \begin{split}
        x_{i(t)} &= x_{i(t-1)} + \Delta x_{i(t)} + \sum_{j \in \mathcal{N}_i} x_{j(t-1)} w_{ji}\\
    	y_{i(t)} &= y_{i(t-1)} + \Delta y_{i(t)}  
    \end{split}
    \label{eq:network}
\end{align}
where 
$\mathcal{N}_i$ is the set of indices of the oscillators connected to the $i$-th oscillator,
$w_{ji}$ is the weight between $i$-th oscillator and $j$-th oscillator.
The connected oscillators in the CPG-based controller cooperate to achieve the desired task.
The number of weights that need to be optimized for the controllers of the robots, spider, gecko, and baby are 18, 13, and 16, respectively.

\paragraph{Steering}
The usual steering policy for wheeled robots is relatively easy, for left (right) turn the force needs to be reduced on the left (right) wheel. Here we generalize this idea to modular robots with no assumptions about the morphology. Our method allows for scaling the activation signals for the joints depending on the coordinates of the joint and the angle $\alpha$ between the direction to the target and the robots heading. The key idea is to use a scaling factor 
\begin{equation}
d_p(\alpha) = \left(\frac{\beta - | \alpha |}{\beta}\right)^p ,
\label{eq:cp}
\end{equation}
where $p > 0$ is a user parameter that determines how strongly we penalize the deviation $\alpha$. 
In this study, we use the value of 7 for the parameter $p$ by the experiments of parameter tuning.
Recall that robots field of vision is the region between $-\beta$ and $\beta$, hence $\alpha \in [-\beta, \beta]$, and $\alpha < 0$ means that the target is on the left, $\alpha > 0$ means the target is on the right.

This scaling factor is used to modify the signals to the joints on the left as follows: 
\begin{equation}
  sig = \begin{cases}
        d_p(\alpha) \cdot out & \textit{if } \alpha < 0 \\
        out  & \textit{if }\alpha \ge 0
    \end{cases}
    \label{eq:l_joint_downscale}
\end{equation}
and, analogously, the signal for the joints on the right is modified as follows:
\begin{equation}
  sig = \begin{cases}
        out & \textit{if } \alpha < 0 \\
        d_p(\alpha) \cdot out  & \textit{if }\alpha \ge 0 .
    \end{cases}
    \label{eq:r_joint_downscale}
\end{equation}
The signals for the middle joints are never modified.


Observe that by these formulas we define a specific implementation of the square shaped extra node within a sensory oscillator in Figure \ref{fig:framework}. We use $d_p(\alpha)$ as the sensory information $sen$ and the function $f$ is simple multiplication. To our knowledge, this is a novel method. The only other existing work that is remotely similar is that of \cite{Ijspeert2007From}, where an internal signal is used to modify the working of oscillators for two predefined locomotions, walking and swimming.

\begin{figure*}[!tbp]
    \centering
    \begin{adjustbox}{max width=0.99\textwidth}
	\begin{tabular}{c c c}
	    \Large{spider} & \Large{gecko} & \Large{baby} \\
	    \includegraphics[width=0.4\textwidth]{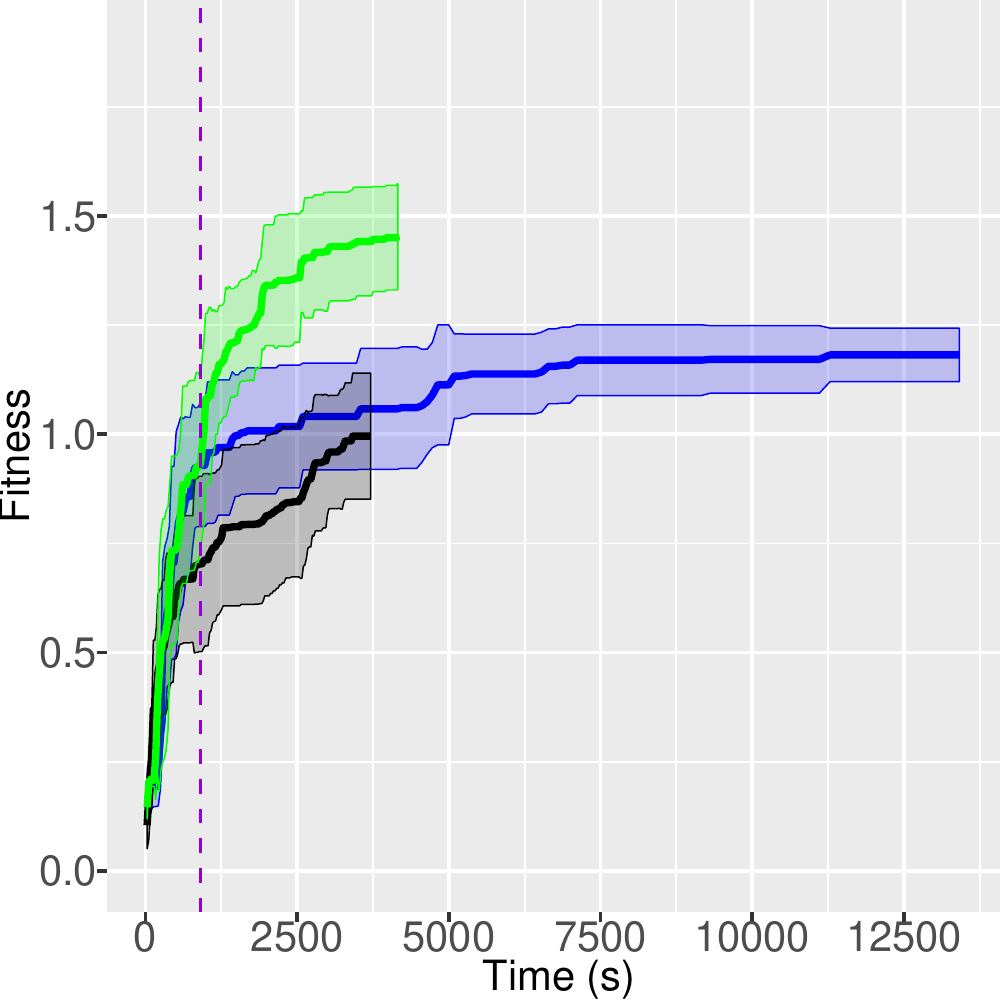} &
		\includegraphics[width=0.4\textwidth]{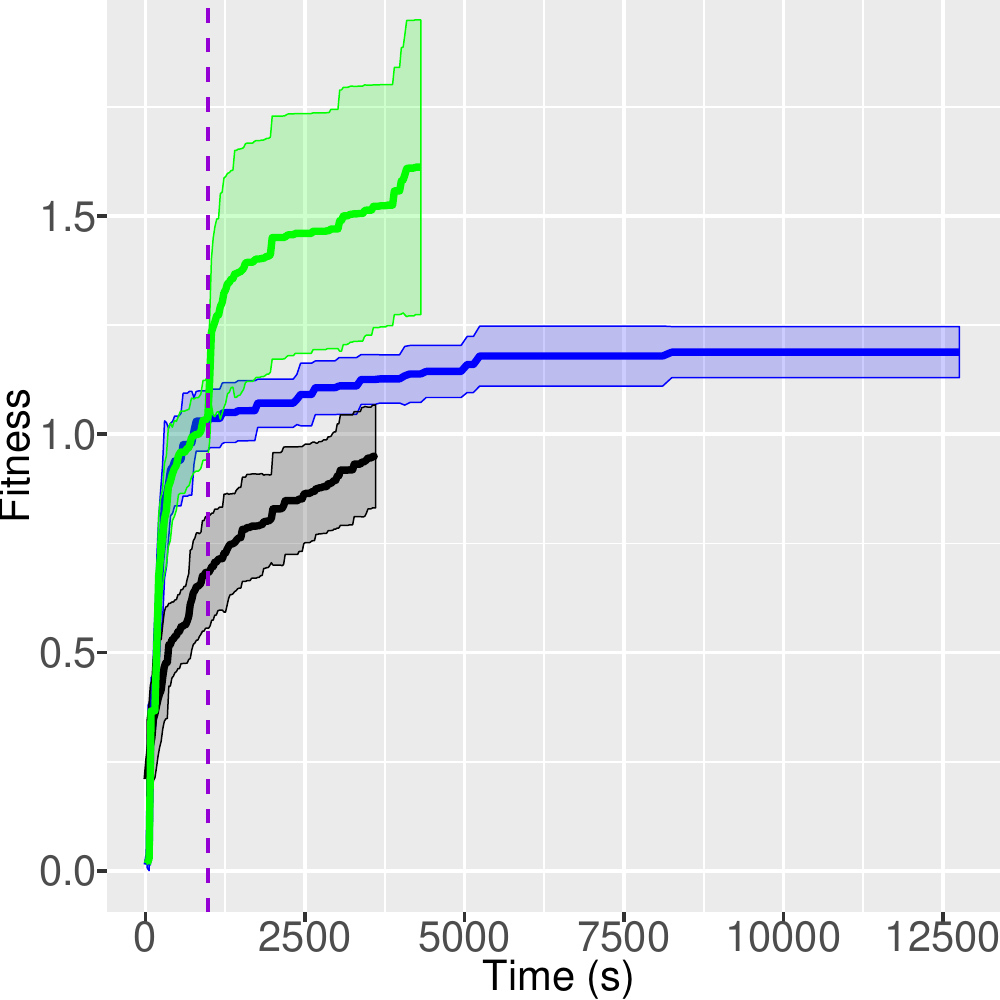} &
		\includegraphics[width=0.4\textwidth]{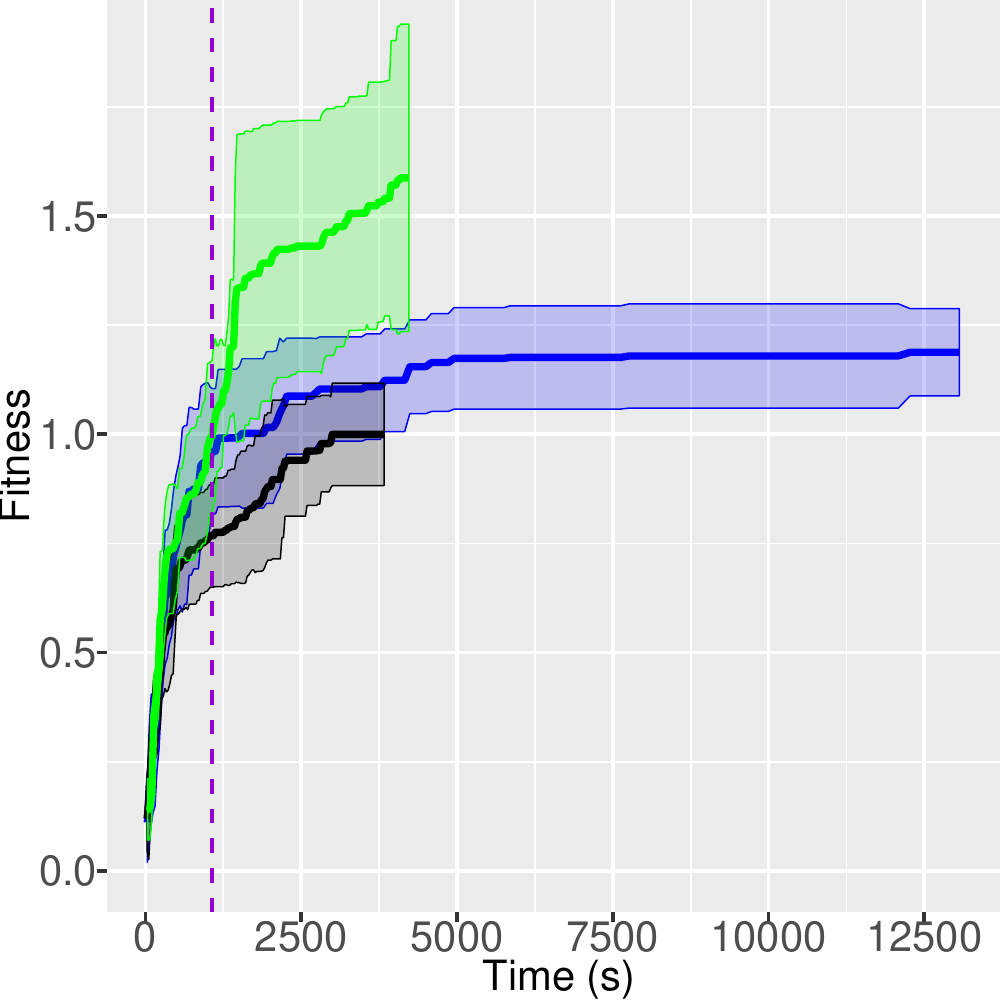} 
	\end{tabular}
    \end{adjustbox}
    \caption{The performance of Bayesian optimization (blue), an evolutionary algorithm (black), and the BEA (green) on the directed locomotion task in simulation. In each plot we provide the mean value (a solid line) with two standard deviations (a shadowed region). The purple dashed lines present the switch points. The plots on the left, in the middle and on the right represent results for spider, gecko, and baby, respectively.}
    \label{fig:bea_comparison}
\end{figure*}

\subsection*{Fitness function for directed locomotion}
\label{sec:function}
%
The fitness function used here is to evaluate the performance of controllers for the task of directed locomotion. This task is defined by a target direction $\gamma$ that the robot has to follow. The good fitness function needs to combine two objectives: minimizing deviation with respect to the target direction $\gamma$ and maximizing speed (i.e., displacement) over the evaluation period (60 seconds in our experiments). 
To calculate the fitness value we need 
\begin{itemize}[noitemsep,nolistsep]
    \item the robots starting position $p_0$ at the beginning of the evaluation period,
    \item the robots final position $p_1$ at the end of the evaluation period,
    \item the deviation angle $\delta$ between the target direction $\gamma$ and the line drawn between $p_0$ and $p_1$,
    \item the total length $L$ of the trajectory travelled during the evaluation period (which is not the distance between $p_0$ and $p_1$).
\end{itemize}  

The fitness value is then composed of several components. Component one is to maximize the distance travelled in the right direction. This distance can be expressed by the value $E_1 = d(p_0,p_1) \times cot(\delta)$, where $d(x,y)$ is the Euclidean distance between two points $x$ and $y$ and $cot$ is the cotangent function. Another component is to minimize the deviation w.r.t. the target angle. This can be expressed not only by $\delta$ but also by the distance of the of the final position $p_1$ from the ideal trajectory starting at $p_0$ and following the target direction. This distance can be expressed as $E_2 = d(p_0,p_1) \times tan(\delta)$, where $tan$ is the tangent function. The third component is to reward locomotion in a straight line. This can be simply captured by the value $E_3 = d(p_0,p_1) / (L + \epsilon$), where $\epsilon$ is an infinitesimal constant. This value is maximized when the length $L$ of the travelled trajectory equals the distance between the starting point $p_0$ and the end point $p_1$. Combining these components into one formula we obtain the following fitness function:

\begin{equation}
\label{eq:fitness}
F = E_3 \cdot \Big{(} \frac{E_1}{\delta + 1} - w \cdot E_2 \Big{)},
\end{equation}
where $w > 0$ is a penalty factor. This function maximized when $d(p_0,p_1)$ is maximal, $\delta$ (and hence $E_2$) is zero, and $L = d(p_0,p_1)$.

The fitness of a given controller in a robot is established by running the robot with that controller, measuring $p_0, p_1, \delta$ and $L$ and calculating the value of $F$ as defined by Equation \ref{eq:fitness}.

\subsection*{The Bayesian-Evolutionary Algorithm}
\label{subsec:BEA}

The Bayesian-Evolutionary Algorithm (BEA) consists of three stages: BO, switching, EA. In the first stage, i.e., the early iterations, Bayesian optimization is employed because its computation time is not yet large. In this paper we use standard Bayesian optimization from the flexible high-performance library Limbo \cite{cully2018limbo}, using a Gaussian process with a Mat\'ern 5/2 kernel, the length scale $\theta = 0.2$, and a GP-UCB acquisition function. This setting outperformed other hyperparameter settings in our preliminary experiments on parameter tuning.

The stage of switching starts when the time efficiency of BO drops to lower than that of EA. To determine the switch point we monitor the quality gain per time interval during the search process. This can be defined for any interval of $n$ consecutive iterations (objective function evaluations). If $t_1, \dots, t_n$ denote the time instances of these iterations and $f_1, \dots, f_n$ the resulting objective function values, then the gain over this time interval is $\mathcal{G} = \frac{f_n - f_1}{t_n - t_1}$. 
We tested the gain of BO and EA on several well-known objective functions and found that a good moment to switch generally lies in the interval between 190 and 300 iterations.
In this study, the switch is triggered at 300 evaluations. 

To seed the EA with a good initial population, we aim for quality (which can be exploited) as well as diversity (which assures appropriate exploration). Hence, if the intended population size of the EA is $K$, then we use K-means clustering in the top $50\%$ of solutions found by BO and transfer the best solution in each cluster to the EA. 

In the third stage, the search is continued by an evolutionary algorithm. In general, this can be any EAs, but here we use an evolution strategy where the mutation step-size is self-adaptive, but is also controlled by the quality gain per time interval. The idea is to use smaller mutation step size for exploitation when the gains are relatively large and larger mutation step sizes for exploration when gains are small over a period of iterations. 
For parameters of the EA in BEA, the mutation rate is 0.8, the population size is 10, the tournament size is 2.

We performed additional experiments to compare the performance of the BEA against its components separately, namely, the BO and the EA. The goal of these additional experiments is to indicate the benefit of our approach compared to using either the BO or the EA alone. We report the performance of the three methods in \autoref{fig:bea_comparison}. Additionally, we present a comparison of the BEA with the BO and the EA in terms of the achieved fitness value and the computational time in \autoref{tab:comparison}. We notice that the BEA obtains around $20-30\%$ better fitness value than the BO and around $45-70\%$ better fitness value than the EA. Obviously, the BEA is slower than the EA (by around $10-20\%$), but it is significantly faster than the BO (by about $70\%$). Eventually, we see that the proposed optimization procedure allows to not only significantly reduce time complexity of the optimization as originally planned, but also leads to a better exploration/exploitation, and, eventually, to better results than the standalone BO.

In conclusion, we want to stress out the novelty of the proposed optimization strategy. First, the idea of combining BO and EAs is not widely-used. Typically, BO is applied to parameter tuning of EAs \cite{karroblack, roman2016bayesian}. Here, we propose to optimize the initial population of the EA using BO. Second, running both algorithms one after the other is not necessarily beneficial. The crucial step is to decide about the moment to switch from a computationally heavy, but accurate procedure, to a lightweight generate-and-test method. Here, we discussed how to accomplish that by monitoring the time efficiency. Third, we propose heuristics to \textit{transfer} solutions found by BO to the initial population of the EA. Last, we further propose a new self-adaptive mutation operation that takes into account information about the progress of the procedure, i.e., the gain in the fitness function value.

\begin{table}[!tbp]
    \centering
    \small
    \setlength{\tabcolsep}{8pt}
    \renewcommand{\arraystretch}{1.2}
    \begin{tabular}{l|l|ccc}
    \multicolumn{2}{c|}{}  &  \multicolumn{1}{c}{spider9} & \multicolumn{1}{c}{gecko7} & \multicolumn{1}{c}{babyA}  \\ 
    \hline
    best fitness ($\uparrow$) & BO & 122.9\% & 135.6\% & 133.6\% \\
    of BEA w.r.t. & EA  & 145.9\% & 169.7\% & 158.7\% \\ 
    \hline
    comp. time  ($\downarrow$) & BO   & 31.0\% &  33.8\% & 32.4\% \\
    of BEA w.r.t. & EA  & 112.2\% & 119.6\% & 110.4\% \\ 
    \end{tabular}
    \caption{Simulation based comparison. Upper half: The performance of BEA in terms of fitness over a full run w.r.t. BO and EA defined by BEA/BO and BEA/EA respectively. Lower half: BEA vs. BO and the EA in terms of computation time defined as BEA/BO and BEA/EA respectively.}
    \label{tab:comparison}
\end{table}

\section*{Data and code availability}
The data and code in this study can be provided by the corresponding author 
upon reasonable request.
A Supplementary Video is available for this paper at \url{https://youtu.be/U9n86ngVe-4}.

\bibliographystyle{unsrt}
\bibliography{Bio-inspired-robots-bibliography}




\end{document}